\documentclass{article}
\usepackage{ijcai15}

\usepackage{times}
\usepackage{graphicx}
\usepackage{url}
\usepackage{epstopdf}
\usepackage{subfigure}
\DeclareGraphicsExtensions{.eps, .png, .jpeg, .jpg}

\title{A PCA-Based Convolutional Network}
\author{Yanhai Gan$^1$, Jun Liu$^1$, Junyu Dong$^2$, Guoqiang Zhong$^2$\\
Ocean University of China\\
Qing Dao, China \\
1:\{gyh5421,liujunqd\}@163.com \\
2:\{dongjunyu,gqzhong\}@ouc.edu.cn}

\begin{document}

\maketitle

\begin{abstract}
  In this paper, we propose a novel \emph{unsupervised} deep learning model, called PCA-based Convolutional Network (PCN). The architecture of PCN is composed of several feature extraction stages and a nonlinear output stage. Particularly, each feature extraction stage includes two layers: a convolutional layer and a feature pooling layer. In the convolutional layer, the filter banks are simply learned by PCA. In the nonlinear output stage, binary hashing is applied. For the higher convolutional layers, the filter banks are learned from the feature maps that were obtained in the previous stage. To test PCN, we conducted extensive experiments on some challenging tasks, including handwritten digits recognition, face recognition and texture classification. The results show that PCN performs competitive with or even better than state-of-the-art deep learning models. More importantly, since there is no back propagation for \emph{supervised} finetuning, PCN is much more efficient than existing deep networks.
\end{abstract}

\section{Introduction}
Traditional models for classification tasks are generally composed of hand-crafted feature extraction and a trainable classifier. The most popular hand-crafted features include Gabor features~\cite{tao2007general}, locally binary patterns (LBP)~\cite{guo2010completed}, Hog~\cite{onishi20083d} and SIFT~\cite{ke2004pca}. They have been successfully applied in texture classification, face recognition and object recognition tasks. However, features extracted by hand-crafted methods are always low-level and suited to specific data and tasks with prior knowledge.

Recently, deep learning has become a popular way of automatically learning features from data that disentangles the underlying factor of variations. The proposed deep approaches always include layerwise stacking of feature extractors. For example, deep belief networks are composed of stacking pre-trained restricted Boltzmann machines (RBMs) and deep auto-encoders are stacked by RBMs or auto-encoders. Deep architectures lead to learn more hierarchical and more abstract features at higher layers of representations.

One of the most powerful deep architectures is a biologically inspired model -- convolutional networks (ConvNets). ConvNets are    a trainable multi-stage architecture with each stage composed of three layers: the filter banks layer, non-linearity layer and feature pooling layer. Weight sharing in the convolution layer and pooling operations are the key of the ConvNets which lead to features invariant to small variations. A deep ConvNets with multistage architecture can learn hierarchical features, from local low-level features to global high-level ones. However, training such a deep network typically uses gradient descent method in a supervised mode, which always need a large scale of labeled samples for training. In addition, good results sometimes depend on the tricks of the trade for parameter tuning, e.g. using "dropout" for regulation ~\cite{abelson-et-al:scheme}.

Recent research has shown that using unsupervised learning in each stage of ConvNets helped reducing the requirement of labeled data significantly. PCANet is such a variation of deep convolutional networks of which convolution filter banks in each stage are simply chosen from PCA filters~\cite{chan:pcanet}. Surprisingly, when such simple filters are used in a deep network architecture, it has demonstrated competitive performance with other deep networks. However, PCANet dispenses with the pooling layer in the feature learning stage, but only uses block-wise histogram together with nonlinear operation in the output stage. This results in the exponentially growing dimensions and training time with increasing number of samples.

In this paper, we propose a convolutional architecture in which the filters are learned from PCA in an unsupervised mode. The network is composed of feature extraction stage which could be stacked to multiple stages and a nonlinearity stage. Feature extraction stage includes a convolution layer and a pooling layer and can be easily cascaded to a deep architecture. The nonlinearity stage includes binary hashing and histogram statistics; the output is then fed into a trainable classifier. The filter bank in convolution layer is learned by PCA, and the generated feature maps are aggregated by pooling layers. This results in multiple sets of feature maps corresponding to different filters which probably detect distinctive features (e.g. detect features at similar orientations) of the input. The filter banks in the higher convolution layer are computed based on combinations of multiple sets of feature maps. This is inspired by the intuition that high level features are the combinations and abstract of low level features. Multiple feature maps corresponding to an input represent different features extracted from the same input. Experiments show the comparative performance in classification tasks against state-of-the-art approaches.
\section{Related Work}
In the past few years, variations of convolutional network have been proposed with respect to the pooling and convolutional operation. Recently, unsupervised learning was used for pre-training in each stage that would alleviate the need of labeled data. When all the stages were pre-trained, the network was fine-tuned by using stochastic gradient descent method. Many methods were proposed to pre-train filter banks of convolution layers in an unsupervised feature learning mode. The convolutional versions of sparse RBMs ~\cite{bgf:Lixto}~\cite{brachman-schmolze:kl-one} , sparse coding~\cite{bruna:invariant} and predictive sparse decomposition(PSD)~\cite{bgf:Lixto}~\cite{gls:hypertrees}~\cite{levesque:functional-foundations}~\cite{levesque:belief} were reported and achieved high accuracy on several benchmarks.

Alternatively, some networks similar to ConvNets were proposed but used pre-fixed filters in convolution layer and yielded good performance on several benchmarks. In  ~\cite{nebel:jair-2000}~\cite{mutch:multiclass}, Gabor filters were used in the first convolution layer. Meanwhile, wavelet scattering networks (ScatNet) ~\cite{bruna:invariant}~\cite{sifre:rotation} also used pre-fixed convolutional filters which were called scattering operators. By using a similar multiple levels of ConvNets, the algorithm had achieved impressive results in the applications of handwritten digits and texture recognition. One more closely related work is called PCANet~\cite{chan:pcanet}, which simply use PCA filters in an unsupervised learning mode at the convolution layer. Built upon a multiple convolution layers, a nonlinear output stage was applied with hashing and block-wise histogram. Just a few cascaded convolution layers were demonstrated to be able to achieve new records in several challenging vision tasks, such as face and handwritten recognition, and comparative results on texture classification and object recognition.

\section{The PCA-Based Convolutional Network}

\begin{figure*}
  \centering
  \includegraphics[width=15cm]{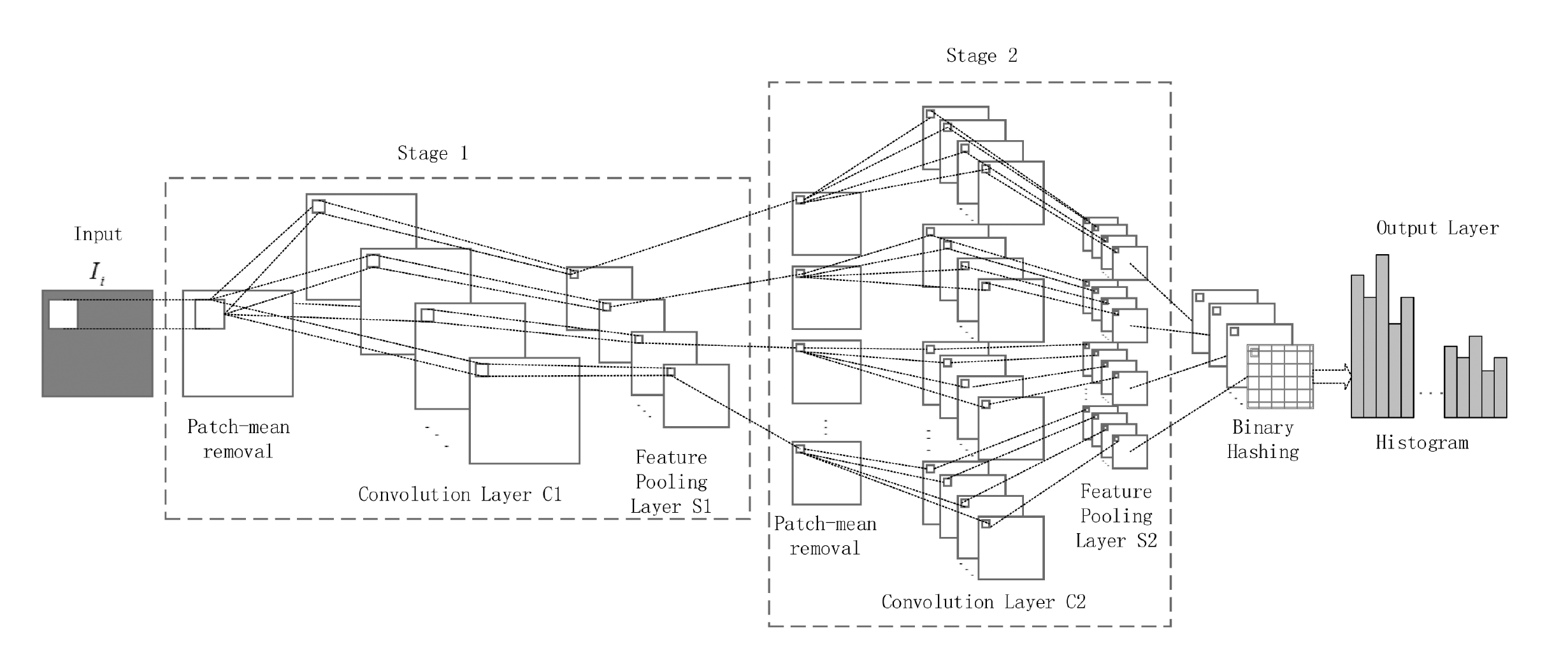}\\
  \caption{The detailed block diagram of the proposed (three-stage) PCN.}\label{fig:img2}
\end{figure*}

 The PCN is essentially a multi-stage convolutional network that can be trained layer-wise in an unsupervised manner. It is composed of cascaded feature extraction stages and a nonlinear output stage. Figure \ref{fig:img2} illustrates the structure of a typical PCN with three stages including the output stage. Each feature extraction stage consists of a convolutional layer and a pooling layer. The inputs are first convoluted with PCA-based filters to produce a set of feature maps.  The pooling layer generally computes the average or maximum value over a neighborhood. The purpose of a pooling layer is to build robustness to small distortions and reduce the resolution of feature maps by a factor $p$ horizontally and $q$ vertically. The propagated feature maps through the pooling layer are then fed into the next stage as input. The final output stage of PCN comprises binary hashing and block-wise histogram statistics.

%\subsection{Our Model}

Suppose we are given $N$ input images which are denoted as $\{\textsl{I}_i\}_{i=1}^{N}$ ; the size of each input image is $m \times n$. The filter size used in each stage is represented as $k_1\times k_2$. In the following we describe each stage of PCN in detail.

\subsection{The first feature extraction stage in PCN}

Inspired by weight sharing of receptive fields in ConvNets~\cite{bgf:Lixto}, for each input image, we sample a number of patches with a size of $k_1\times k_2$ at every $k$ pixel locations, i.e. the sample interval is $k$ pixels.  Each patch is vectorized to form a column with $k_1k_2$ elements. Then all patches sampled from the same input image are put together to form a matrix of size $(k_1k_2)\times((\lceil\frac{m-k_1}{k}\rceil+1)(\lceil\frac{n-k_2}{k}\rceil+1))$, denoted as $X_i=[x_{i,1},x_{i,2},x_{i,3},\cdots,x_{i,(\lceil\frac{m-k_1}{k}\rceil+1)(\lceil\frac{n-k_2}{k}\rceil+1)}]\in R^{(k_1k_2)\times((\lceil\frac{m-k_1}{k}\rceil+1)(\lceil\frac{n-k_2}{k}\rceil+1))}$ , where $x_{i,j}$ represents the vector of the $jth$ patch in $\textit{I}_i$.

In order to introduce competitions between adjacent features within a neighbourhood, each column vector in the matrix $X_i$ subtracts the mean value of the corresponding patch to obtain the matrix $\bar{X}_i=[\bar{x}_{i,1},\bar{x}_{i,2},\bar{x}_{i,3},\cdots,\bar{x}_{i,(\lceil\frac{m-k_1}{k}\rceil+1)(\lceil\frac{n-k_2}{k}\rceil+1)}]$ . This operation is reminiscent to the local contrast normalization used by ImageNet~\cite{krizhevsky2012imagenet}. Once matrices for all input images are constructed in the same way, they are assembled to form a large matrix  $X=[\bar{X}_1,\bar{X}_2,\bar{X}_3,\cdots,\bar{X}_N]$ . Subsequently, each row of $X$ subtracts its mean, the result is also denoted as $X$.
Eigenvalue decomposition is then performed on the matrix $XX^T$.  The convolutional filters are selected as the first $L_1$ principle eigenvectors of $XX^T$. Thus, the learned filters can be described as $W_l=mat_{k_1,k_2}(q_l(XX^T))\in R^{k_1\times k_2},l=1,2,3,\cdots,L_1$, where $mat_{k_1,k_2}(v)$  denote the mapping relationship from vector $v$ to a matrix $W\in R^{k_1 \times k_2}$, $q_l(XX^T)$ represent the $lth$ eigenvector of matrix $XX^T$. The eigenvectors are reshaped to the size $k_1\times k_2$. In this way, we obtain $L_1$ filters of size $k_1\times k_2$. We subsequently convolute the learned filters with the input images to generate filter responses at each pixel location; we call the filtering results feature maps. \begin{equation}\label{eq1}
  I_i^l=I_i*W_l,i=1,2,3,\cdots,N
\end{equation}
where: $*$ is 2D convolution operation; $I_i$ is padded with zeros before convolution.

The convolution with each input image produces $L_1$ feature maps. Each feature map represents particular features extracted at corresponding location in the image.  We divide the feature maps (padded with zeros) generated by the convolutional layer into several non-overlapping pooling regions of size $p\times q$.  Then the max pooling or average pooling is applied to the pooling regions. The pooling operation results in feature maps with reduced resolution, and these pooling features are robust to small distortions. We use $S_i^l$ to represent the pooling result of the $lth$ feature map of the $ith$ input image. Given a  collection $\{I_i\}_{i=1}^N$ of $N$ input images, through the convolution and pooling operation using the $lth$ filter we obtain N feature maps, which are denoted as $\{S_i^l\}_{i=1}^N,l=1,2,3,\cdots,L_1$. Since there are $L_1$ filters in the first extraction stage, we obtain $NL_1$ feature maps in total.
\subsection{The second feature extraction stage in PCN}

\begin{figure}
  \centering
  \includegraphics[width=8.5cm]{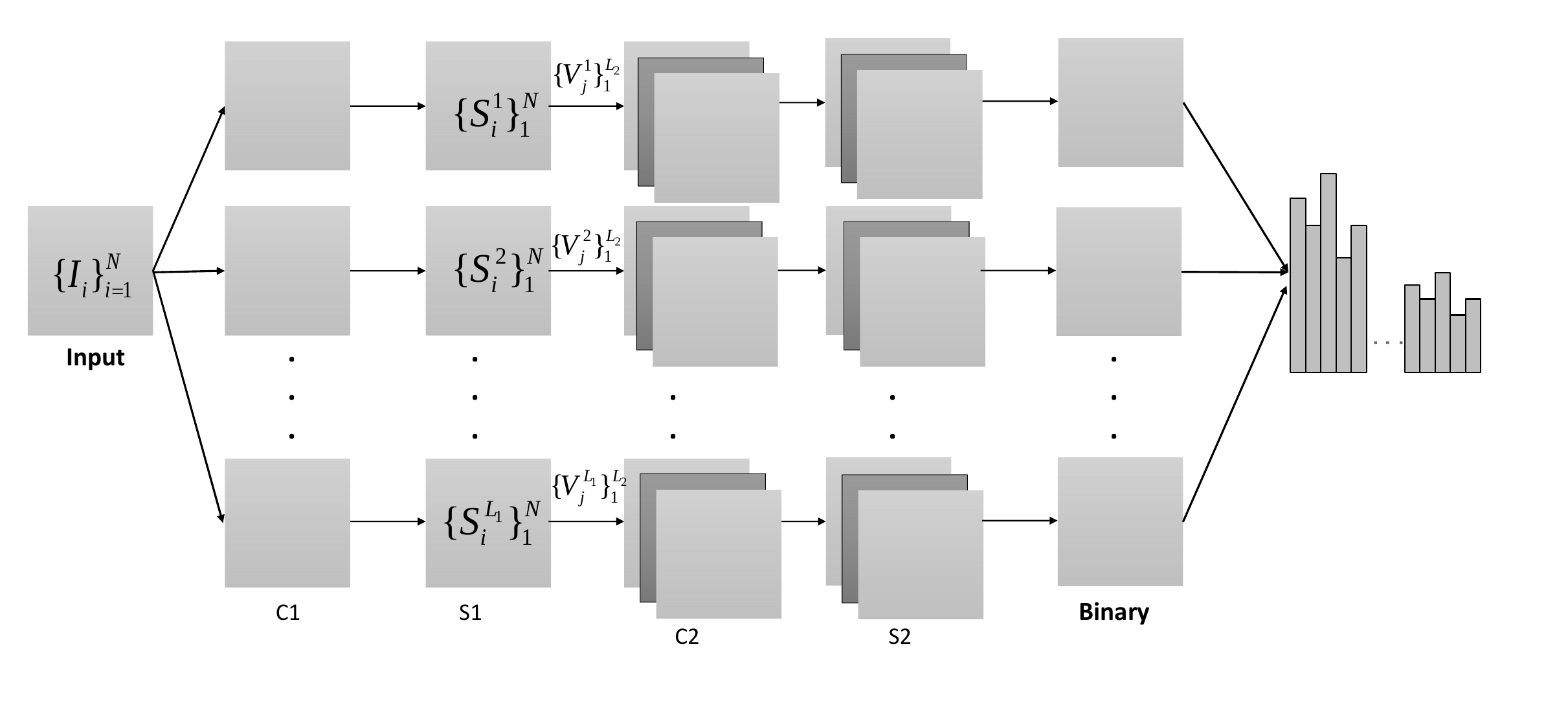}\\
  \caption{Basic structure of the proposed (three-stage) PCN.}\label{fig:img3}
\end{figure}
The pooled feature maps in the first stage are treated as the original input to the second stage. These $NL_1$ feature maps are divided into $L_1$ subsets. Each subset includes $N$ feature maps which are produced by convoluting the input images with the same filter in the previous stage, and they are denoted as $S^l=\{S_i^l\}_{i=1}^N,l=1,2,3,\cdots,L_1$.  Feature maps in one subset capture certain features of the input images, whereas those in different subsets capture different types of features. Figure \ref{fig:img3} shows the structure of the proposed PCN.

Since high level features are the combinations and abstract of low level features~\cite{gutmann2013three}, we combine subsets $\{S^l\}_{l=1}^{L_1}$ according to certain rule to form several groups. Table 1 demonstrates one way to combine the subsets. In each group(corresponding to a column in the table), feature maps(marked with ¡±$\times$¡±) corresponding to the same input image are added. The combined subsets are then used as the actual input to the second feature extraction stage.

\begin{table}
  \centering
  \caption{An example of combination ways. The first layer consist of 5 filters;  two adjacent subsets of feature maps are combined£©.}
  \label{tab1}
  \begin{tabular}{lccccc}
  \hline
  &1&2&3&4&5\\ \hline
  $S^1$ & $\times$ & & & & $\times$\\ \hline
  $S^2$ & $\times$ & $\times$ & & &\\ \hline
  $S^3$ & & $\times$ & $\times$ & &\\ \hline
  $S^4$ & & & $\times$ & $\times$ &\\ \hline
  $S^5$ & & & & $\times$ & $\times$\\ \hline
  \end{tabular}

\end{table}
In Table \ref{tab1}, each row represents a subset of feature maps obtained from the previous stage, and each column represents a group combining these subsets in a particular way. There are 5 filters in the first stage which result in 5 subsets. Two of the subsets are added and 5 groups are formed. In the table, ¡±$\times$¡± indicates corresponding subsets are combined to form one group. In practice, an indexing matrix is used to define the way of combination. In the indexing matrix, most entries are zeros and a few of entries are ones, which indicate the subsets belonging to one group. Thus, different indexing matrices can be defined. If the indexing matrix is defined as an identity matrix, there will be no combination of subsets.

The combination produces several new subsets and each new subset is denoted as $\{S_i^{l'}\}_{i=1}^N$ , which also consists of $N$ feature maps.

By repeating the same procedure as in the first stage, for each $\{S_i^{l'}\}_{i=1}^N$, we sample patches from each feature map in this subset. Then we also subtract patch mean values and join all vectors together to form a matrix denoted as $\bar{Y}_i^{l'}=[\bar{y}_{l',i,1},\bar{y}_{l',i,2},\bar{y}_{l',i,3},\cdots,\bar{y}_{l',i,(\lceil\frac{m-k_1}{k}\rceil+1)(\lceil\frac{n-k_2}{k}\rceil+1)}]$, where $\bar{y}_{l',i,j}$ represents the mean removed vector of the $jth$ patch of the $ith$ feature map in the $l'th$ subset. We further collect patches from all the feature maps in this subset, remove the patch mean, and concatenate the matrixes $\bar{Y}_i^{l'}$ as $Y^{l'}=[\bar{Y}_1^{l'},\bar{Y}_2^{l'},\bar{Y}_3^{l'},\cdots,\bar{Y}_N^{l'}]$. Afterwards the row mean is removed form $Y^{l'}$. Since there are $L_1$ subsets, we obtain $L_1$ such matrixes $Y^{l'},l'=1,2,3,\cdots,L_1$.\\\\
For each subset $\{S_i^{l'}\}_{i=1}^N$, we construct filters using the following equation separately:
\begin{equation}\label{eq2}
  V_l^{l'}=mat_{k_1,k_2}(q_l(Y^{l'}{Y^{l'}}^T))\in R^{k_1\times k_2},l=1,2,3,\cdots,L_2
\end{equation}
For each subset, we choose the first $L_2$ principle eigenvectors as PCA filters, denoted as $\{V_l^{l'}\}_{l=1}^{L_2}$. Each input feature map in this subset is convoluted with $L_2$ filters, which resulted in $L_2$ new feature maps. Since there are $L_1$ subsets(produced by $L_1$ groups), we produce $L_1NL_2$ feature maps in total in the second stage, and they are the output of the second feature extraction stage (C2 in Figure \ref{fig:img3}).

The pooling process in the second stage is the same as in the first stage. The output feature maps of C2 are divided into several non-overlapping patches with size $p\times q$ and the maximum or average value is calculated over the pooling region.

If there are more feature extraction stages, the process is repeated in the same way as described above.

\subsection{The output stage in PCN}
In the output stage we reconstruct feature maps to form final representations of the  input image.  We use binary hashing and histogram statistics (called "hashingHist") as in PCANet~\cite{chan:pcanet}. Each input feature map $S_i^{l'}$ to the second stage produces $L_2$ output maps. We binarize these output maps and calculate $H(S_i^{l'}*V_l)$, where $H(.)$ is a Heaviside step function whose value is one for positive entries and zero otherwise. For each pixel location, we treat the vector of $L_2$ binary bits as a decimal number. This converts the $L_2$ outputs generated in the second stage back into a single integer-valued ¡°image¡±.

For each of the $L_1$ integer-valued images, we partition it into $B$ blocks. We compute the histogram of the decimal values in each block, and concatenate all the $B$ histograms into one vector. After this encoding process, the ¡°feature¡± of the input image $I_i$ becomes the set of block-wise histograms. The local blocks can be either overlapping or non-overlapping, depending on applications.

\section{Experiments}
In all experiments, a three-stage (including the final output stage) PCN is applied to different data sets for simplicity. The final output features of the PCN are sent to a linear SVM for classification. All these configurations keep fixed. We compared the efficiency of PCN for different recognition tasks using the same desktop PC with an Intel i5-3570 CPU and 32GB memory.

\subsection{Digit Recognition based on the MNIST Datasets}
Because images in the MNIST Datasets are small, we set the patching sampling interval as 1, i.e. we sample a patch at each pixel location. The patch size is set as $7\times7$. In the output stage, we set the block size as $7\times 7$, and we set the block overlapping ratio as 0.5. The three parameters keep unchanged during the experiment. In particular the pooling layer is disabled in every feature extraction stage, and it can be easily controlled by a parameter in our code. We select an identity matrix as the indexing matrix, that is, we make every group in the second stage contain only one subset.

\subsubsection{Digit recognition on the basic MNIST Dataset}
%\begin{figure}
%  \centering
%  \includegraphics[width=8.5cm]{6.eps}\\
%  \caption{Accuracy varies as the filters in the second layer increase(patch size is set to 7, step is 1 and the filters in the first layer is 6).}\label{fig:img6}
%\end{figure}
%\begin{figure}
%  \centering
%  \includegraphics[width=8.5cm]{7.eps}\\
%  \caption{Accuracy varies as the filters in the first layer increase(patch size is set to 7, step is 1 and the filters in the second layer is 10).}\label{fig:img7}
%\end{figure}
%\begin{figure}
%\begin{minipage}[t]{0.48\linewidth}
%\centerline{\includegraphics[width=1.5in]{7.pdf}}
%\centerline{a}
%\end{minipage}
%\hfill
%\begin{minipage}[t]{0.48\linewidth}
%\centerline{\includegraphics[width=1.5in]{6.pdf}}
%\centerline{b}
%\end{minipage}%
%\caption{Accuracy varies with the filters in the first stage(a) and second stage(b).}
%\label{fig:img6}
%\end{figure}

\begin{table}
  \centering
  \caption{Comparison of digit recognition rates(\%) of different methods on Basic MNIST.}
  \label{tab3}
  \begin{tabular}{lc}
  \hline
  Method&Accuracy\\ \hline \hline
  PCANet-2 & 98.94  \\ \hline
  CAE-2 & 97.52  \\ \hline
  ScatNet-2 & 98.73  \\ \hline
  PCN-2 & \textbf{99.20} \\ \hline
  \end{tabular}

\end{table}
%\begin{figure}
%  \centering
%  % Requires \usepackage{graphicx}
%  \includegraphics[width=8.5cm]{12.eps}\\
%  \caption{Filters in the first stage on Basic MNist.}\label{fig:img12}
%\end{figure}
%\begin{figure}
%  \centering
%  % Requires \usepackage{graphicx}
%  \includegraphics[width=8.5cm]{13.eps}\\
%  \caption{Filters in the second stage on Basic MNist(Filters of each group are shown as a row).}\label{fig:img13}
%\end{figure}

 The basic MNIST dataset is a smaller subset of MNIST. It contains 10000 training images, 2000 validation images and 50000 testing images. We first perform our experiment on the basic dataset. The hyper-parameters were selected to maximize the performance on the validation set. Then, the system was trained over the entire training set and validation set. We achieve the highest accuracy of 99.20\% when the numbers of filters in the first stage and second stage are set to 6 and 11 respectively.This is higher than related methods in literature.

% Figure \ref{fig:img4}a illustrates the relationship between the accuracy and the number of filters in the first layer. Here the number of filters in the second stage is set to 10. We can find that when the number of filters in the first stage increases to 8, the accuracy can reach the peak of 99.10\%. In conclusion, the accuracy increases as the number of filters increases whether in the first or second stage in the primary time, and reaches its peak when the number of filters becomes large enough, and then the accuracy begins to decline as the number keeps increasing. We find that a appropriate configuration of the number of filters in every stage is important to classification accuracy. We find the highest accuracy of 99.20\% in this experiment when the numbers of filters in the first stage and second stage are set to 6 and 11 respectively.
\subsubsection{Digit recognition on the standard MNIST Dataset}
The standard MNIST dataset consists of 60000 training images and 10000 testing images. To adjust hyperparameters, a validation set of 5 samples per class was taken out of the training sets. The hyper-parameters were selected to maximize the performance on the validation set. Then, the system was trained over the entire training set. We found the best configuration when the numbers of filters in the first and second stage were set to 8 and 10 respectively, and the accuracy reached 99.41\%, which outperformed the related works, as shown in table \ref{tab4}.
Overall, PCN can achieve competitive performance compared to the state-of-the-art, but with much less computation due to its simple network structure.
\begin{table}
  \centering
  \caption{Comparison of digit recognition rates(\%) of different methods on standard MNIST.}
  \label{tab4}
  \begin{tabular}{lc}
  \hline
  Methods & Accuracy\\ \hline \hline
  HSC~\cite{learning:sparse} & 99.23  \\
  K-NN-SCM~\cite{belongie2002shape} & 99.37  \\
  K-NN-IDM~\cite{keysers2007deformation} & 99.46  \\
  CDBN~\cite{lee2009convolutional} & 99.18  \\
  ConvNet~\cite{bgf:Lixto} & 99.47  \\
  ScatNet-2($SVM_{rbf}$)~\cite{bruna:invariant} & \textbf{99.57}  \\
  PCANet-2 & 99.34  \\ \hline
  PCN-2 & \textbf{99.41} \\ \hline
  \end{tabular}

\end{table}

\subsection{Face Recognition on the Extended Yale B Dataset}
\begin{figure}
  \centering
  \includegraphics[width=8.5cm]{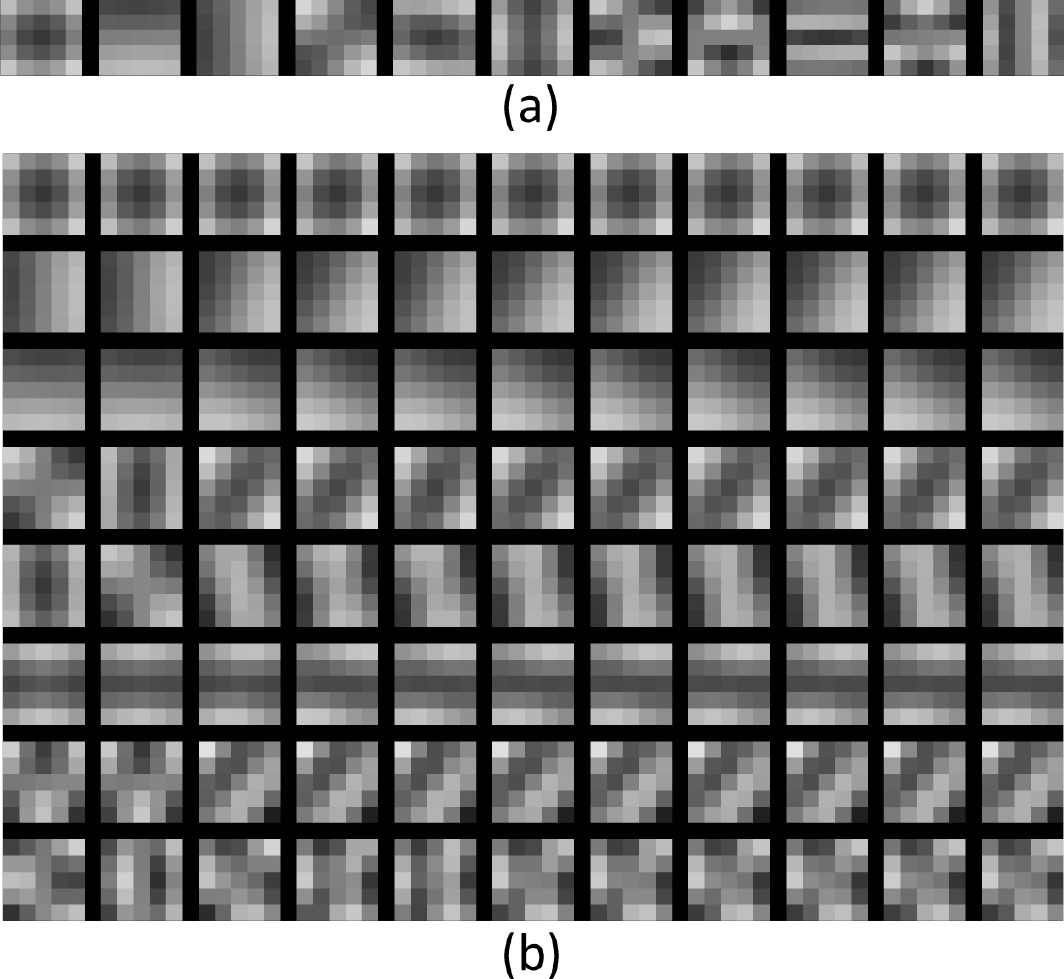}\\
  \caption{Filters learned on the Extended Yale B dataset. (a) 11 filters in the first stage;(b) There are 11 groups in the second stage, and each group contains 8 filters, shown in a column.}\label{fig:img8}
\end{figure}

\begin{table}
  \centering
  \caption{Face recognition rates(\%) and time consumption(s) on Extended Yale B.}
  \label{tab5}
  \begin{tabular}{lcc}
  \hline
  Methods & PCANet-2 & PCN-2\\ \hline \hline
  Accuracy & 99.53 & 99.58  \\
  Training Time & 8551.75 & 2054.42  \\
  Test Time/Sample & 1.39 & 0.27  \\ \hline
  \end{tabular}

\end{table}
The extended Yale B dataset contains 2414 frontal-face images of 38 individuals. The cropped and normalized $192\times168$ face images were captured under various lighting conditions. For each subject, we randomly select 5 images as our testing images, and the rest for training. A validation set of 5 images per subject was taken out of the training sets. The hyper-parameters were selected to maximize the performance on the validation set. Then, the system was trained over the entire training set. In the end the patch size was set as $5\times5$, and the numbers of filters in the first and second stage were set as 11 and 8 respectively. The patch sampling interval was set as 1. The max pooling module used a $2\times2$ boxcar filter with a $2\times2$ down-sampling step. We used non-overlapping blocks in the output stage and the block size was set as $8\times8$. Identity matrix was used as the indexing matrix in the second stage. We achieve the average accuracy of 99.58\% over 10 experiments, as shown in table \ref{tab5}. The training time of our method including PCN plus SVM is $2054.42s$, and the testing time per sample is 0.27. This is much more efficient compared to PCANet. The filters in the first stage are shown in figure \ref{fig:img8}a; it is obvious that each filter in the first stage captures direction-related features of an input face image. Each column in Figure \ref{fig:img8}b contains filters in one group in the second stage; it can be seen that the filter banks in different groups are similar to a large extend, but there are still some differences, so we can't use just the same filters in different groups. We found that an identity matrix was better than other matrix when used as the indexing matrix, this may be caused by the blur effect that all subsets in on group connected to the same filter bank, so we maybe use different filters in the future.

%\begin{figure}
%  \centering
%  \includegraphics[width=8.5cm]{9.eps}\\
%  \caption{Filters in the second stage on Extended Yale B(Filters of each group are shown as a column).}\label{fig:img9}
%\end{figure}

%\begin{table}
%  \centering
%  \begin{tabular}{lc}
%  \hline
%  Methods & Accuracy & Train Time & Test Time\\ \hline \hline
%  PCANet-2 & 99.58  \\ \hline
%  PCN-2 & \textbf{99.58}  \\ \hline
%  \end{tabular}
%  \caption{Face recognition rates(\%) and time consumption(s) on Extended Yale B.}\label{tab5}
%\end{table}
%& \textbf{2054.42} & \textbf{0.27} & Training Time(s) & Testing Time(s/sample)
\subsection{Texture Classification on CUReT Dataset}
The CUReT texture dataset contains 61 categories of textures. Each category contains images of the same material with different pose and illumination conditions. In this experiment, following PCANet~\cite{chan:pcanet}, a subset of the original data with azimuthal viewing angles less than 60 degrees was selected, thereby yielding 92 images in each class. A central $200\times200$ region was cropped from each of the selected images. The dataset was randomly split into a training and a testing set, with 46 training images for each class. The hyper-parameters were selected according to literature. The filter size was set as $5\times 5$; the patch sampling interval was set as 1. The number of filters in both stage was set as 8, and non-overlapping block size was $50\times 50$. The pooling layer was disabled in each extraction stage. Identity matrix was used as indexing matrix in the second stage. The accuracy reached 99.71\%, which was higher than the result of 99.61\% achieved by PCANet.
\begin{table}
  \centering
  \caption{Comparison of accuracies(\%) on CUReT dataset.£©}
  \label{tab6}
  \begin{tabular}{lc}
  \hline
  Methods & Accuracy\\ \hline \hline
  Textons & 98.50  \\
  BIF & 98.60  \\
  Histogram & 99.00  \\
  ScatNet-2(PCA) & \textbf{99.80}  \\
  PCANet-2 & 99.61  \\ \hline
  PCN-2 & \textbf{99.71} \\ \hline
  \end{tabular}

\end{table}
\subsection{Texture Classification on Outex Dataset}
Outex is a framework for empirical evaluation of texture classification and segmentation algorithms. Problems are encapsulated into welldefined test suites having precise specifications of input and output data. Outex database contains surface textures and natural scenes. The collection of surface textures is expanding continuously. At this very moment the database contains 320 surface textures, both macrotextures and microtextures. Many textures have variations in local color content, which results in challenging local gray scale variations in intensity images. Some of the source textures have a large tactile dimension, which can induce considerable local gray scale distortions. Each source texture is imaged according to certain procedure. The images used in a texture classification suite are extracted from the given set of source images (particular texture classes, illuminations, spatial resolutions, and rotation angles) by centering the sampling grid so that equally many pixels are left over on each side of the sampling grid. If the training and testing images of a particular texture classification problem are extracted from the same set of source images, the images are divided randomly to two halves of equal size for the purpose of obtaining an unbiased performance estimate. The directory images in each test suite includes the images needed in the test suite. The directory indexed by three numbers includes the specified problem in this test suite. Each one of these directories has three files:classes.txt,test.txt and train.txt which define the problem.  The problem indexed by $000$ of the $Outex_TC_00004$ test suite was selected in our experiment.
After several validating trails, the patch size was set as $5\times5$. The patch sampling interval was set as 1, and the numbers of filters in the first and second stage were set as 18 and 6 respectively. The pooling layer was disabled in each stage. The block size was set as $14\times14$ and the block overlapping ratio was 0.5. An identity matrix is used as the indexing matrix in the second stage. We achieve the classification accuracy of 99.91\%, and the training time is $260.84s$ including PCN and SVM. What's more the test time per sample is $0.15s$.
%\begin{figure}
%  \centering
%  \includegraphics[width=8.5cm]{10.eps}\\
%  \caption{Filters in the first stage on Outex.}\label{fig:img10}
%\end{figure}
%\begin{figure}
%  \centering
%  \includegraphics[width=8.5cm]{11.eps}\\
%  \caption{Filters in the second stage on Outex(Each column corresponds to a group).}\label{fig:img11}
%\end{figure}

\subsection{Texture Classification on Our Dataset}
\begin{figure}
 \centering
 \includegraphics[width=8.5cm]{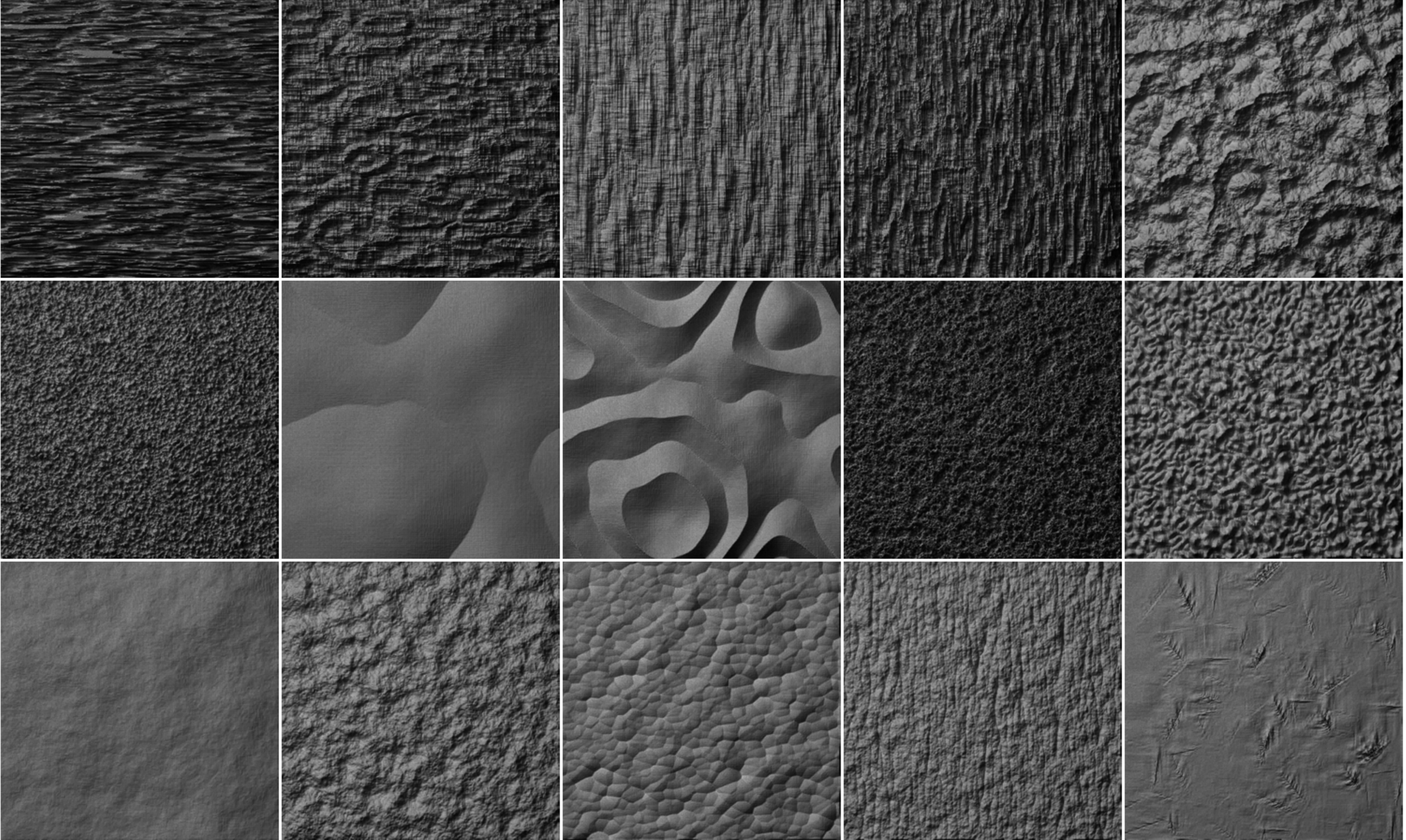}\\
 \caption{Example samples of our texture dataset.}\label{fig:img4}
\end{figure}

Procedural models are widely used in computer graphics for generating realistic, natural-looking textures. A number of procedural models have been proposed and these models can produce various textures. Through render these textures are presented as surface images. Given a surface image, it is important to know which model can produce such kind of texture.  This is  a typical texture classification problem. Our procedural texture dataset contains a number of rendered textures generated by 23 procedural texture models and then rendered by Luxrender given fixed light conditions. Textures generated by one method normally are different from those generated by other methods; however, some textures produced by different models may be perceived similar. This forms a challenging classification task. Figure \ref{fig:img4} shows example samples of our texture dataset.

\begin{figure*}
  \centering
  \includegraphics[width=15cm]{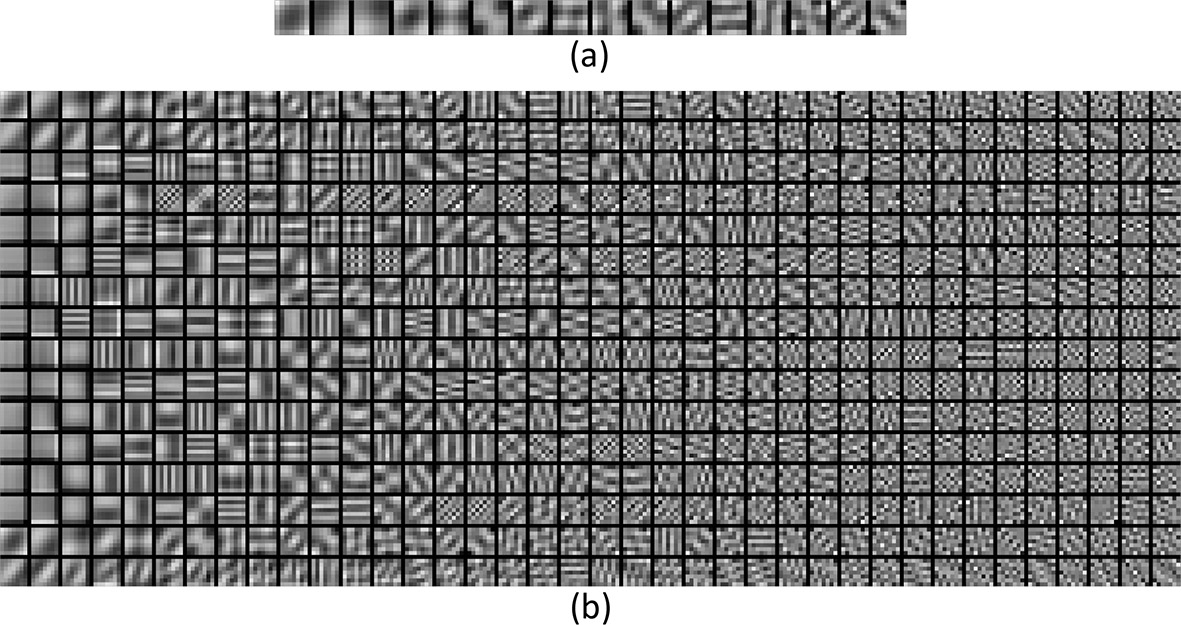}\\
  \caption{Filters at the first(a) and second stage(b) on our texture dataset. There are 16 filters in the first stage. There are 16 groups in the second stage and every group contains 38 filters, which are shown in a row.}
  \label{fig:img14}
\end{figure*}

The size of surface images in our dataset is 256*256. In this experiment, we use a total of 3600 surface images, which will be available together with the source code in the near future.

We randomly choose 25\% of the images from each method as our testing set and the rest are used for training.  A validation set of 5 samples per method was taken out of the training set. The hyper-parameters were selected to maximize the performance on the validation set. We found the best configuration that patch size $7\times7$, patch sampling interval 3, the numbers of filters in both extraction stages $L_1=16,L_2=38$. A $2\times2$ boxcar filter with a $2\times2$ down-sampling step was used in the pooling layer. In particular the output non-linear stage was removed. All feature maps from the feature extraction phase were reshaped and concatenated to form a vector as the input to the linear SVM classifier. Then the PCN was trained over the entire training set using the best configuration. The accuracy reaches 99.89\% which is higher than the result of 99.62\% achieved by PCANet. These results are shown in Table \ref{tab2}. More importantly, our algorithm is much more efficient than PCANet in terms of computation cost. The large numbers of filters suggest that surface images in our dataset contains complex structure.

Figure \ref{fig:img14}(a) shows that most filters in the first stage extract orientation related features of input images. Due to the complex structure of our texture, some filters look complicated. A fact is that some surface images have no obvious edge information. In figure \ref{fig:img14}(b), each row contains filters of one group in the second stage. It can be observed that prior filters in a group can extract large scale features, and posterior filters extract more detailed features.

As a comparison, we also use a traditional CNN for the same classification task with the same computation resources. After running 10 hours for 50000 iterations, we only achieve an accuracy of 43.2\%. The performance becomes worse as the number of iterations increases. It is obvious that the CNN falls into overfitting because we do not have enough training samples.

\begin{table}
  \centering
  \caption{Comparisons of different methods on our texture dataset.£©}
  \label{tab2}
  \begin{tabular}{lcc}
  \hline
  &PCN-2&PCANet-2\\ \hline
  Accuracy(\%) & 99.89 & 99.62 \\ \hline
  Train time(s) & 251.80 & 16407.50\\ \hline
  Test time per sample(s) & 0.1136 & 3.14 \\ \hline
  \end{tabular}

\end{table}

%\subsection{Object Recognition on Cifar10}
%We finally evaluate the performance of PCN on CIFAR10 dataset for object recognition. The CIFAR-10 dataset consists of 60000 $32\times32$ colour images in 10 classes, with 6000 images per class. There are 50000 training images and 10000 test images. Images in CIFAR10 vary significantly not only in object position and object scale within each class, but also in colors and textures of these objects. To begin with, we extend PCN filter learning so as to accommodate the RGB images in object datasets. In the same spirit of constructing the data matrix X in (1), we gather the same individual matrix for RGB channels of the images, denoted by Xr, Xg, Xb ¡Ê Rk1k2¡ÁNmn, respectively. Following the key steps in Section 2.1.1, the multichannel PCA filters can be easily verified as

\section{Conclusion}
We propose a PCA-based Convolutional Network (PCN), which essentially has the advantage of both CNN~\cite{bgf:Lixto} and PCANet~\cite{chan:pcanet}, i.e. it can achieve competitive performance compared with state-of-the-art methods but is much more efficient in terms of computation. The PCN used in our experiments simply comprises two feature extraction stages and a non-linearity output stage. However, instead of training the network by using iteration methods, PCN simply uses PCA to learn filters in convolution layer. The eigenvectors are used as the filters to convolute with the input images.

Similar to other deep networks, it should be noted that a proper configuration of PCN is very important for different types of inputs. If training images are relatively simple in terms of structure and have a large size, we can use a relatively large interval to sample the patches and enable the pooling layer to rapidly reduce the feature dimension of the input image. On the other hand, if the input image is small enough, we may simply set the patch sampling interval to one and disable the pooling layer. In the grouping process, all subsets in one group are connected to the same filter bank, so we can add up all the subsets to form a new subset. But different filter banks maybe work more effectively. We consider to use different filter banks in one group in the future.

\section*{Acknowledgment}
This Work Was Supported By National Natural Science Foundation Of China(NSFC) (No. 61271405) ;
The Ph.D. Program Foundation Of Ministry Of Education Of China (No. 20120132110018);

\newpage
%\section*{Acknowledgment}
%This Work Was Supported By National Natural Science Foundation Of China(NSFC) (No. 61271405) ;
%The Ph.D. Program Foundation Of Ministry Of Education Of China (No. 20120132110018);
\bibliographystyle{named}

\end{document}